\documentclass[runningheads]{llncs}
\usepackage{graphicx}

\usepackage{tikz}
\usepackage{comment}
\usepackage{amsmath,amssymb} 
\usepackage{color}

\usepackage[accsupp]{axessibility}  

\usepackage{multirow}

\usepackage{xcolor}
\newcommand{\ie}{\emph{i.e.,}\xspace}
\newcommand{\eg}{\emph{e.g.,}\xspace}

\newcommand{\etal}{\emph{et~al.}\xspace}
\usepackage{subfig}
\usepackage[colorlinks=true,citecolor=blue]{hyperref}
\usepackage[T1]{fontenc}

\usepackage{amsmath}
\usepackage{booktabs}
\usepackage{graphicx}
\usepackage{amsfonts}

\usepackage{algorithm}

\newcommand{\Inp}{\textbf{Input:}\hspace*{0.5em}}
\newcommand{\Out}{\textbf{Output:}\hspace*{0.5em}}

\newcommand{\X}{\hspace*{3mm}}
\newcommand{\XX}{\X\X}

\newcommand{\cm}[1]{$\triangleright$ #1}
\usepackage{array}
\newcolumntype{C}[1]{>{\centering\arraybackslash}p{#1}}

\usepackage{bbm}
\usepackage{bm}
\usepackage{xspace} 

\begin{document}
\pagestyle{headings}
\mainmatter
\def\ECCVSubNumber{629}  

\title{Towards Open Set Video Anomaly Detection} 

\titlerunning{Towards Open Set Video Anomaly Detection}
%
\author{Yuansheng Zhu
\and
Wentao Bao
\and
Qi Yu
}
\authorrunning{Zhu, Bao, and Yu}
%
\institute{
Rochester Institute of Technology \\
\email{\{yz7008, wb6219 and qi.yu\}@rit.edu}}
\maketitle

\begin{abstract}
   Open Set Video Anomaly Detection (OpenVAD) aims to identify abnormal events from video data where both known anomalies and novel ones exist in testing. Unsupervised models learned solely from normal videos are applicable to any testing anomalies but suffer from a high false positive rate. In contrast, weakly supervised methods are effective in detecting known anomalies but could fail in an open world. We develop a novel weakly supervised method for the OpenVAD problem by integrating evidential deep learning (EDL) and normalizing flows (NFs) into a multiple instance learning (MIL) framework. Specifically, we propose to use graph neural networks and triplet loss to learn discriminative features for training the EDL classifier, where the EDL is capable of identifying the unknown anomalies by quantifying the uncertainty. Moreover, we develop an uncertainty-aware selection strategy to obtain clean anomaly instances and a NFs module to generate the pseudo anomalies.
   Our method is superior to existing approaches by inheriting the advantages of both the unsupervised NFs and the weakly-supervised MIL framework. Experimental results on multiple real-world video datasets show the effectiveness of our method.
\end{abstract}

\section{Introduction}

\begin{figure*}[t]
\centering
\includegraphics[width=0.95\linewidth]{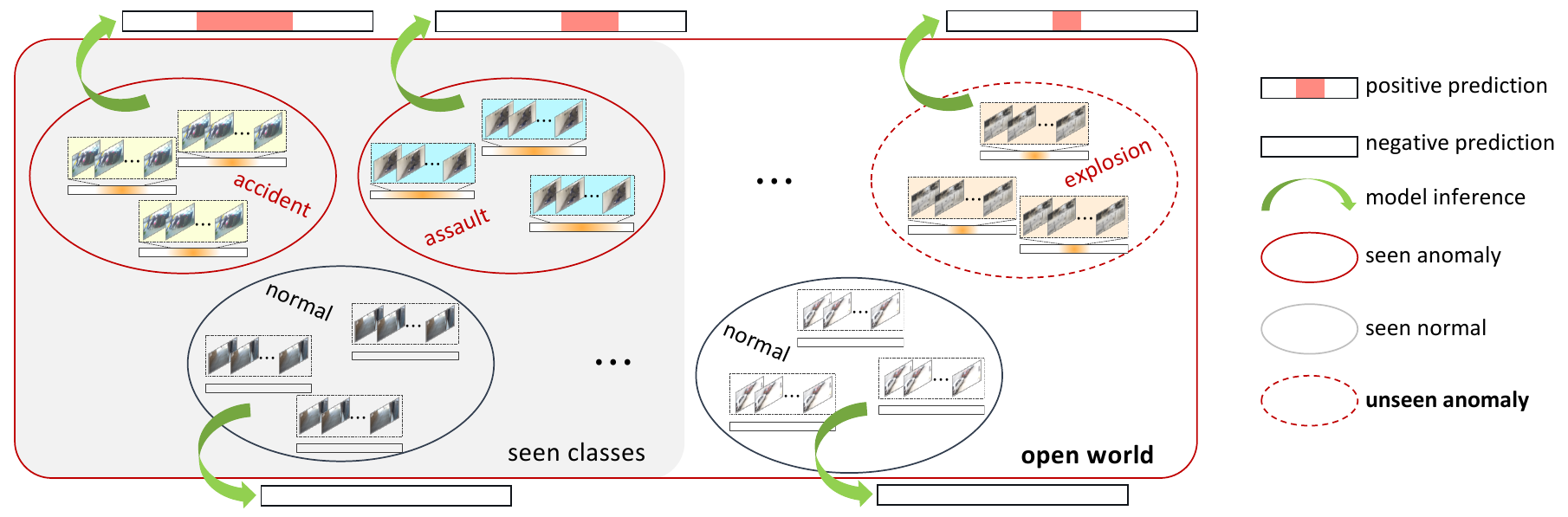}
\caption{\small\textbf{OpenVAD Task.} We propose to conduct OpenVAD through the weakly-supervised video anomaly detection in an open world. In training, only a closed-set of anomaly videos with video-level annotations are observed along with normal videos. In testing, the model is asked to identify and localize the anomaly events from videos in an open world where arbitrary unseen anomalies could exist.
\label{fig:problem}
}
\end{figure*}

Traditional video anomaly detection aims to detect abnormal events that significantly deviate from normal ones. Examples of such abnormal events include human crimes, natural disasters, and traffic accidents, to name a few. It has been successfully applied to many real-world applications~\cite{ramachandra2020survey}. However, unseen novel anomalies may occur after a well-trained supervised model has been deployed in an open world. Being aware of the unseen anomalies essentially leads to the Open-set Video Anomaly Detection (OpenVAD) problem (see Figure~\ref{fig:problem}), which is under-explored in literature despite being critical to real-world applications.

Unsupervised learning is one of the typical ways to handle unseen anomalies in existing literature~\cite{liu2018future,lu2013abnormal,tran2017anomaly,yu2020cloze,hasan2016learning}. They aim to learn representative features of normal events only from normal videos. However, as unsupervised methods neglect the anomaly information, they are in general less effective to detect complicated anomalies that are similar to normal samples in the feature space~\cite{borisyak20191+}. Besides, (weakly-)supervised methods can use annotated anomaly videos so that their performances are generally better. However, the learned models are limited to detecting a closed set of anomaly categories and unable to handle the arbitrary unseen anomalies. Inspired by the recent advances in open set recognition~\cite{ScheirerTPAMI2012,bendale2016towards,du2022vos,Bao_2021_ICCV}, we propose to perform video anomaly detection in an open world 
by extending the existing weakly-supervised video anomaly detection paradigm. The goal is to achieve accurate anomaly detection in an open set by learning a model only with video-level annotations from a closed subset of anomaly types. 

Most existing works~\cite{sultani2018real,zhong2019graph,wu2020not,tian2021weakly} formulate weakly supervised anomaly detection as a multiple instance learning (MIL) problem~\cite{dietterich1997solving}, where a video is modeled as a bag of instances (\eg video clips). Due to the lack of fine-grained instance labels, simply assigning all instances in an abnormal (\ie positive) bag with the same anomaly label inevitably incurs severe labeling noise. Such noise further results in a high false positive rate, \ie falsely treating normal events as abnormal ones, when using a trained model for detection. Therefore, it is essential to select the clean anomaly instances in MIL. Besides, to enable the MIL model to be aware of the unseen anomaly in testing, the open space risk of MIL should be bounded while it is under-explored in the existing literature.

To tackle these challenges, we first leverage graph neural networks (GCNs) and a triplet loss to learn representative instance features. Then, we instantiate the MIL with evidential deep learning (EDL)~\cite{sensoy2018evidential,shi2020multifaceted,Bao_2021_ICCV} and use the predicted evidence to help select anomaly instances with high \emph{cleanness}
for robust MIL training. 
To bound the open space risk, an unsupervised normalizing flows (NFs) is learned from normal instances, which allows us to sample pseudo anomaly samples from the low density area of the instance distribution.

We are among the first few attempts (\eg \cite{liu2019margin}) that  address video anomaly detection in the open set setting, where arbitrary unseen anomaly events could appear in testing. This is fundamentally more challenging than a using fully-supervised training process as adopted by some existing efforts in~\cite{borisyak20191+,ryzhikov2019normalizing} and some of these methods are not suitable to tackle high-dimensional video data. Though there are few works~\cite{kopuklu2021driver,doshi2022rethinking} with a similar open set assumption, our task definition is more realistic in real-world, where the model is weakly-supervised and aims to identify arbitrary unknown anomaly
events in testing. In summary, the key contributions are threefold: 
\begin{itemize}
    \item We formulate a novel MIL framework for the OpenVAD problem to detect both seen and unseen video anomalies in a challenging open-world setting. 
    \item We integrate MIL with a normalizing flow-based generative model and evidential learning for high-quality instance selection.
    \item We conduct extensive experiments and the results show significant superiority to existing unsupervised and weak supervised learning approaches.   
\end{itemize}

\section{Related Work}

\paragraph{Video Anomaly Detection.} Existing methods to detect anomalies from videos can be categorized into two groups: 1) unsupervised learning and 2) weakly supervised learning. Unsupervised approaches are motivated by the premise that anomaly events are fundamentally challenging to completely enumerate and characterize. Therefore, these approaches conduct either dictionary learning/sparse coding \cite{lu2013abnormal,zhao2011online} or construct a deep auto-encoder~\cite{yang2015unsupervised,chen2015detecting,tran2017anomaly}. 
Unlike unsupervised approaches, weakly supervised approaches~\cite{liu2019margin,sultani2018real,he2018anomaly,zhong2019graph,wu2020not,tian2021weakly} use both normal videos and anomaly videos for training. They take advantage of normal videos and leverage additional anomaly videos to train a binary classifier over the video clips.
Among existing weakly supervised approaches, one representative approach leverages a deep auto-encoder to locate clips in anomaly videos with low reconstruction scores and then further reduces these scores through margin learning~\cite{liu2019margin}. Some other approaches~\cite{sultani2018real,he2018anomaly,zhong2019graph} formulate this problem as multiple instance learning (MIL) problem and build an instance-level classifier. To train the classifier properly, these MIL models select one or a set of instances from a positive bag to feed into the classifier and either a fixed number of instances are selected~\cite{sultani2018real,he2018anomaly} or a predefined threshold is set as the selection criterion~\cite{zhong2019graph}. Previous MIL based approaches achieve superior performance compared to unsupervised counterparts. The core design of these approaches lies in the measurement of \emph{cleanness} and the instance selection strategy. In this paper, we empirically show that the top-$k$ strategy is robust, which inspires us to develop the \emph{cleanness} measurement and an instance selection strategy by introducing the evidence based learning method.

\paragraph{Open Set Recognition.} Open Set Recognition (OSR) aims to identify the unknowns while keeping reasonable closed set performance in the open set testing.
Since the unknowns are never seen by the model and are out-of-distribution (OOD) with respect to the in-distributional (ID) training data, they are referred to as ``unknown unknown class" in existing literature~\cite{GengTPAMI2020,yang2021generalized}. In contrast, the exact classes of ID data are known and are called ``known known classes". The OSR approaches attempt to classify test samples from ``known known classes" and detect test samples from ``unknown unknown classes". The most recent work~\cite{perera2019deep} encourages learning a multi-class classifier to give low activation values for external OOD data and large values for ID data. Some works do not depend on external OOD data in training by using either specialized loss functions~\cite{Bao_2021_ICCV,charpentier2020posterior} or unknown class generation~\cite{zhou2021learning}. Other works~\cite{zhang2020hybrid,bendale2016towards} address the OSR problem by integrating probabilistic density estimation in the latent feature space with a multi-class classifier. Our proposed OpenVAD and the traditional OSR problems can both be regarded as a generalized OOD detection problem~\cite{yang2021generalized}, because of the semantic distribution shift between testing and training datasets. A key distinction of the proposed OpenVAD is that it does not care about if an anomaly video is from seen or unseen categories in testing. Instead, it focuses on distinguishing arbitrary anomaly activities from normal ones in an open world. 

\section{Methodology}

Our work aims to handle the OpenVAD problem under a weakly supervised setting. Specifically, a model is trained using normal videos and a closed set of untrimmed anomaly videos with video (or bag)-level labels. Each anomaly video contains at least one anomaly event (\ie positive instance). The model is tested in an open world, where unseen anomaly events may occur, and the model is expected to robustly identify all anomaly clips (\ie segments of a video) during testing. Since we only have access to the video-level anomaly class labels, MIL provides an ideal learning framework to leverage the weakly supervised learning signals. 
Specifically, an untrimmed video is regarded as a bag $X$, which contains $N$ temporal clips as instances in the bag:  $X=\{\mathbf{x}_{1}, \ldots, \mathbf{x}_N\}$. The learning objective of MIL is to train an instance-level binary classifier $\Phi: \mathbf{x}_n\rightarrow y_n$ with only bag-level class label $Y\in\{0, 1\}$ in the training, while the instance-level labels $y_n\in\{0, 1\}$ with $y_n\in\{y_{1}, \ldots, y_N\}$ are not available. In MIL, a positive bag $(X,Y=1)$ contains at least one positive instance (\eg anomaly event), \ie $\exists n \in [1,N], y_{n}=1$, while a negative bag consists of only negative instances (\eg normal event), \ie $\forall n\in [1,N],y_{n}=0$.

For the OpenVAD problem, the model is expected to handle the semantically shifted testing data that contains unseen anomaly events. In testing, given an untrimmed video, the model is expected to identify the anomaly clips, which could be either seen or unseen anomaly if the video contains any anomaly events.

\begin{figure*}[t]
\centering
\includegraphics[width=0.95\linewidth]{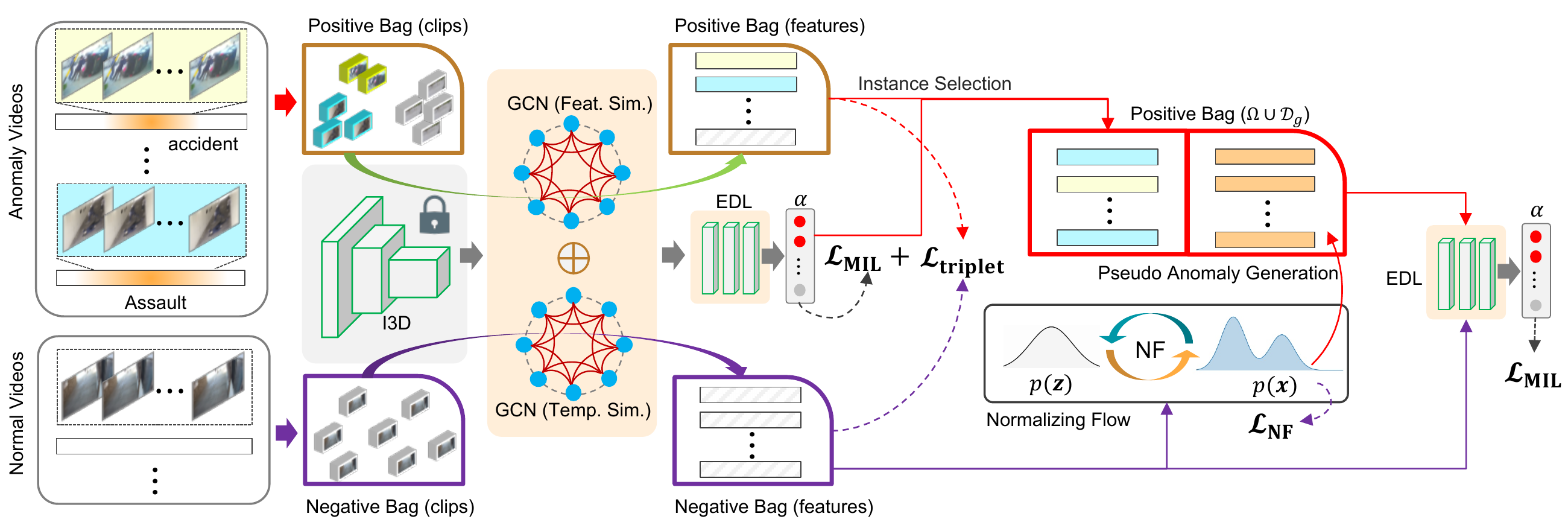}
\caption{ Overview of the OpenVAD Framework. We first integrate two GCNs feature extractors and the evidential deep learning (EDL) into the multiple instance learning (MIL), which are learned by $\mathcal{L}_{\text{MIL}}+\mathcal{L}_{\text{triplet}}$. Then, the classification confidence and uncertainty from the trained EDL head are used to select instances with high \emph{cleanness}. We further utilize the GCNs features of normal video clips to train a normalizing flow model, from which pseudo anomaly features are generated. Together with the normal videos, the selected clean anomaly and the generated pseudo anomaly are gathered to refine the EDL head by $\mathcal{L}_{\text{MIL}}$. 
\label{fig:framework}}
\vspace{-4mm}
\end{figure*}

\paragraph{Overview of the Framework.} 
Our method is developed as a MIL framework, as shown in Figure~\ref{fig:framework}. The instances are structured as graph data to address the temporal consistency and feature similarity of video clips. The MIL framework consists of two graph convolutional networks (GCN) followed by an evidential deep learning (EDL) head, which are trained by weakly-supervised MIL loss and triplet loss. The classification confidence and uncertainty are used to select clean anomaly instances for robust MIL training. To reduce the open space risk in OpenVAD tasks, we further utilize normal videos to train a normalizing flow (NF) model to generate pseudo anomalies. The selected and generated anomalies are used to fine-tune the EDL classification head. In testing, our model could correctly identify an unseen anomaly event from untrimmed video data. See Appendix Section A for the summary of notations.

\subsection{Instance Feature Representation Learning}

We take advantage of a pre-trained action recognition backbone (\eg I3D~\cite{carreira2017quo}) to extract features of raw video clips, denoted as $\mathbf{x}$.
Given the temporal and visual similarity of close clips in a video, we propose to enhance the feature $\mathbf{x}$ by constructing temporal and feature similarity graphs, respectively. Furthermore, to enlarge the feature discrepancy between normal and anomaly instances, a triplet learning constraint is introduced in the feature embedding space.  

More specifically, since each bag is formed from a video, there exist natural dependencies among instances in the bag. Thus, video clips that are located closely in temporal and feature space should have similar feature representation. We leverage a graph neural network to model such dependency. Given a graph defined by the adjacency matrix $A$ and graph node features $X$ (\ie a bag), the graph neural network is updated according to the following rule~\cite{kipf2016semi}:
\begin{equation}
    \mathcal{H}(X, A) = \sigma\left( \hat{D}^{-\frac{1}{2}}\hat{A}\hat{D}^{-\frac{1}{2}}XW\right),
\end{equation}
where $\hat{A} = A + I$ and $I$ is the identity matrix, $\hat{D}$ is the degree matrix of $\hat{A}$, and $W$ is the learnable weights. We employ two GCNs where the adjacent matrix $A$ is determined to capture feature similarity and temporal consistency, respectively~\cite{wu2020not}. The two GCN embeddings are further concatenated to form the final instance representation. To avoid a cluttered notation, we denote the refined instance representation as $\mathcal{H}(\mathbf{x})$.

The GCNs help to compactly embed the neighbouring instances in the same bag. When multiple bags are involved, we further employ a triplet loss to achieve good separability of instances from different classes and compactness of instances in the same class. A triplet consists of an anchor instance, a paired instance from the same class as the anchor, and an unpaired instance from a different class from the anchor, which are sampled from the collection $\Omega$ (defined in Eq~\eqref{eq:selection}). The triplet loss is given by 
\begin{equation}
    \mathcal{L}_{triplet} = \left[d_{ap} - d_{an} + m\right]_{+}
\label{eq:triplet}
\end{equation}
where $m$ is the margin that controls the desired difference between the paired distance $d_{ap}$ and unpaired distance $d_{an}$. 

In addition to GCNs, the consistency loss function~\cite{sultani2018real,tian2021weakly} and other network architectures~\cite{tian2021weakly} (\eg LSTM and CNN) also have been investigated to model the relationship among instances. We choose GCNs because they have been proved to be effective in close set video anomaly detection tasks~\cite{zhong2019graph,wu2020not}. However, the representation learning in the open world is more complex than in a close set. 
The triplet loss will make the normal data stay close to each other while remaining dissimilar to the seen anomaly. This facilitating a normalizing flow (NF) to learn their distribution in the following stage.

\subsection{Uncertainty-aware Multiple Instance Learning}

To train an instance-level classifier with only bag-level annotations, a naive solution is to assign a noisy version of label $y_n$ for each instance $\mathbf{x}_n$ by directly using the bag-level label, \ie $y_n=Y$. As a result, most of instances in the positive bag may be incorrectly labelled (due to normal video clips in anomaly videos) during model training. These mislabeled instances may degrade the model performance as the learning targets are noisy. Furthermore, a MIL model instantiated by deep neural networks (DNNs) is usually sensitive to noisy labels because DNNs are of high capacity to fit random labels~\cite{zhang2021understanding}. Therefore, instance selection from positive bags poses a critical issue that may significantly impact the performance of a MIL model~\cite{wu2020not,zhong2019graph}.

To address this challenge, top-$k$ methods (\eg \cite{7299056}) select the ``most positive" instances from the positive bag in training. 
Two recent works~\cite{tian2021weakly,wu2020not} adapt this technique and achieve decent performance in close set video anomaly detection tasks.  Most top-$k$ methods follow a ``sort-then-rank" process, where a critical step is to determine the \emph{cleanness} of each instance to evaluate how likely the instance is a clean positive one (\ie an anomaly event). In this work, we propose to quantify the classification uncertainty by evidential deep learning. We then introduce an instance selection approach based on uncertainty and confidence.

Given an instance $\mathbf{x}$, we introduce a binary random variable $p\in\{0,1\}$ that represents the chance that it is an anomaly, which follows a Bernoulli distribution. Similar to EDL~\cite{sensoy2018evidential}, we further introduce a Beta distribution $\text{Beta}(\boldsymbol{\alpha})$ as the conjugate prior of the Bernoulli likelihood. In our framework, the learning objective is to train a DNN to directly predict the Beta distribution parameter $\boldsymbol{\alpha}$ instead of $p$. According to Subjective Logic and evidence theory~\cite{JosangBook2016,SentzBook2002}, the predicted parameters $\boldsymbol{\alpha}=(\alpha_+,\alpha_{-})^\top$ can be regarded as the evidences over the positive (anomaly) and negative (normal) classes,  which can be expressed as 
\begin{equation}
    \boldsymbol\alpha = \Phi (\mathcal{H}(\mathbf{x})) + {\bf a} W
\label{eq:alpha}
\end{equation}
where ${\bf a}=(a_+, a_-)^\top$ is the base rate with $a_+= a_-=1/2$ and $W=2$ in our binary case; $\mathcal{H}$ and $\Phi$ denote GCNs based backbone as the feature extractor and a DNN as the EDL classification head, respectively. The GCNs encode the relationship among instances in a bag and output discriminative video clip features. In practice, the EDL classification head consists of two fully connected layers and ReLU activation to predict non-negative class-wise logits. The predictions $\alpha_{+}$ and $\alpha_{-}$ correspond to the instance class $y=1$ and $y=0$ respectively, indicating the virtual quantity of evidence to support $\mathbf{x}$ to be classified as positive or negative. 

The benefit of placing the Beta prior on the Bernoulli likelihood is that the prediction uncertainty can be efficiently and accurately quantified according to recent Dirichlet-based models~\cite{kopetzki2021evaluating}, enabling the MIL model to \emph{know what it does not know}. The predicted probabilities are the ratios of the evidence, \ie  $\mathbb{E}[p_+]=\alpha_+/\alpha_0$ and $\mathbb{E}[p_{-}]=\alpha_-/\alpha_0$, where $\alpha_0=\alpha_++\alpha_-$, and the evidential uncertainty (or vacuity) is quantified as $u=2/\alpha_0$.

Based on the evidential uncertainty above, we propose to train the MIL by selecting the {\em clean} samples, which form a set: 
\begin{align}
\label{eq:selection}
    \Omega = \left\{\mathbf{x}_i|p_+^{(i)} \geq \tau_{p},\alpha_{+}^{(i)} \geq \tau_{u}\right\},
\end{align}
where $p^{(i)}_+$ and $\alpha^{(i)}_{+}$ are the confidence score and evidence that support the sample $\mathbf{x}_i$ being classified as an anomaly; $\tau_{p}$ and $\tau_{u}$ are thresholds to control the size of the set. 
The MIL model is trained by minimizing the Type II maximum likelihood loss (MLL)~\cite{sensoy2018evidential}:
\begin{equation}
    \mathcal{L}_{\text{MIL}}(\mathbf{x}^\prime_i) = \sum_{k=1}^2\left[\hat{y}^{(i)}_{k}\left(\log(\alpha_0^{(i)}) - \log(\alpha^{(i)}_k)\right)\right]
\end{equation}
where the training sample $\mathbf{x}_i^\prime$ is from either the selected set $\Omega$ or the negative bag $\mathcal{N}=\{\mathbf{x}_i|y_i=0\}$, \ie $\mathbf{x}_i^\prime\in \Omega \cup \mathcal{N}$, $\hat{y}$ is the one-hot version of the label $y$. The benefit of Type II MLL lies in that it not only enforces the predicted probability to be as close to the ground-truth bag label, but also increases the evidence for observed data to support better instance selection.

MIL is intrinsically learning using the noisy labels, which is a long-standing challenge. Recent research approaches this problem generally from sample selection~\cite{malach2017decoupling,yu2019does,zaheer2020self,xia2021sample}, sample re-weighting~\cite{wang2017multiclass,chang2017active}, and meta-learning~\cite{li2017learning,zhang2020distilling} perspectives. We choose to explore sample selection because it is well-justified and empirically works well. The sample selection methods often leverage the memorization effect of DNNs to identify the clean samples, which stems from the observation that DNNs tend to fit clean samples faster than noisy ones~\cite{arpit2017closer}. Here, $\tau_{p}$ in Eq.~\eqref{eq:selection} is derived from the memorization effect, which encourages $\Omega$ to retain those qualified samples, \ie samples whose predicted probability $p_{+}$ is most close to the learning target, $y=1$. However, sample selection methods could severally suffer from the bias confirmation~\cite{tarvainen2017mean} caused by incorrect selection, which is challenging to solve. The proposed uncertainty aware MIL is distinguished from  exiting efforts~\cite{wu2020not,wan2020weakly,sultani2018real} for mitigating the bias confirmation by filtering out fluctuating samples (\ie those with a small $\alpha_{+}$ less than $\tau_{u}$), which may be incorrectly selected if only $p_{+}$ is considered.

\subsection{Pseudo Anomaly Generation} 

For the OpenVAD problem, open space risk should be bounded in training so that the model could be aware of the semantically shifted unseen data. However, unseen anomaly videos are not available during training under the open set setting. Inspired by recent generative open set modeling~\cite{du2022vos,article} and normalizing flow, we propose to introduce the normalizing flow to generate pseudo anomaly instances in the latent feature space.  

NFs are the flow-based deep generative models that could explicitly learn the probability density of data~\cite{tabak2013family,tabak2010density}. It has been shown effective for realistic data generation, exact posterior inference, and exact density estimation. NFs consist of composition of invertible learnable functions $f_l$ to transform between a simple distribution $p(\mathbf{z})$ and the complex data distribution $p(\mathbf{x})$, \ie $\mathbf{z}=f(\mathbf{x})$ and $\mathbf{x}=f^{-1}(\mathbf{z})$ where $f=f_1\circ \cdots \circ  f_L$. To enable exact density estimation and data generation, the invertible transformation is constructed following the change-of-variable rule:
\begin{equation}
    p(\mathbf{x}) = p(\mathbf{z})\left|\text{det}\left(\frac{\partial f(\mathbf{x})}{\partial \mathbf{x}^\top}\right)\right|
\label{eq:fx}
\end{equation}
where $\mathbf{z}$ typically follows a non-informative Gaussian distribution. Eq.~\eqref{eq:fx} is associated with the determinant of Jacobian, which can be efficiently computed when a proper bijective function $f_l$ is designed to make the Jacobian triangular. Recent normalizing flow models such as RealNVP~\cite{dinh2016density}, Glow~\cite{kingma2018glow}, and IAF~\cite{kingma2016improved} use different coupling layers to address the functional invertibility and computational efficiency of the transformation $f_l$ based on deep neural networks.

In this paper, we adopt the IAF model because it is efficient for sampling purpose, which is critical for pseudo anomaly generation. The coupling layer $f_l$ is designed as
\begin{equation}
    \mathbf{x}_i = \mathbf{z}_i \odot s(\mathbf{z}_{1:i-1}) + t(\mathbf{z}_{1:i-1})
\end{equation}
where $\mathbf{x}_i$ and $\mathbf{z}_i$ are the $i$-th entries of feature vector $\mathbf{x}$ and $\mathbf{z}$, respectively. The scaling and shifting functions $s$ and $t$ are instantiated by deep neural networks.

To train the flow model, the goal is to maximize the log-likelihood of the normal video data, which is equivalent to minimizing the following objective by stochastic gradient descent:
\begin{equation}
    \mathcal{L}_{\text{NF}} = \mathbb{E}_{\mathcal{D}(X,Y=0)}\left[-\log p(\mathcal{H}(\mathbf{x}))\right]
\end{equation}
Note that here we learn the flow model $f$ in the latent feature space $\mathcal{H}(\mathbf{x})$. This alleviates the difficulty of directly learning in a high-dimensional space.

\noindent\textbf{Pseudo anomaly generation.} The generation is based on the assumption that pseudo anomalies are outliers with respect to the normal class. Therefore, the pseudo anomaly instance $\mathbf{\tilde{x}}$ should be sampled from the low density region of the probability distribution of the normal class $p(\mathbf{x}|y=0)$. In this paper, we first use the tractable $p(\mathbf{z})$ to repeatedly sample the latent feature $\mathbf{\tilde{z}}_i$. Then, the pseudo samples are generated by the flow model of the normal class, \ie $\mathbf{\tilde{x}}=g(\mathbf{\tilde{z}};y=0)$ where $g(\cdot)=f^{-1}(\cdot)$. Finally, only the pseudo samples with low density (\ie with a small $p(\mathbf{\tilde{x}}|y=0)$) are preserved as pseudo anomaly instances. The following equations summarize this procedure:
\begin{equation}
\label{eq:sampling}
\begin{aligned}
     \mathbf{\tilde{z}} \sim p(\mathbf{z}), \;
     \mathbf{\tilde{x}}=g(\mathbf{\tilde{z}};y=0), \;
     \mathcal{D}_g = \{\mathbf{\tilde{x}}|p(\mathbf{\tilde{x}}|y=0) \leq \epsilon \}
\end{aligned}
\end{equation}
where $p(\mathbf{\tilde{x}}|y=0) = p(\mathbf{z})\left|\text{det}\left(\frac{\partial f(\mathbf{x})}{\partial \mathbf{x}^T}\right)\right|$. The benefit of generating pseudo anomaly instances from normal class distribution is that the class boundary of the normal class is constrained by the pseudo anomalies so that the open space risk is well managed. Regarding the way of generating pseudo anomaly instances, unlike previous work~\cite{article} that generates in the pixel space, we perform generation in the feature space, similar to~\cite{du2022vos}. Spawning from the feature space has dual advantages over spawning from the pixel space: 1) the latent space has a much lower dimension, which allows fast sampling for the NFs, and 2) the latent features are more discriminative than the raw pixels~\cite{cho2022unsupervised}; thus the generated data are less likely to be negatively impacted by the background noise in the video data. In contrast, \cite{du2022vos} uses the Gaussian mixtures to model the class-wise in-distribution data density. We argue that the Gaussian distribution is not flexible to model the complex in-distribution data. NFs can seamlessly solve this problem owing to its high capability via a deep composition of the bejiection function $f$.

\noindent\textbf{Discussion. } NFs have a high capability of density estimation while free of supervision. A natural question that arises is whether we can directly ensemble the prediction results of a normalizing flow and a classifier. Ensemble is widely used yet seems to be simpler than pseudo anomaly generation. However, we remark that the ensembles can not inherit the benefit from unsupervised learning and (weakly-) supervised learning paradigms. This is because that NFs cannot detect the complicated anomaly (\eg anomaly samples that look similar to normal ones), which is consistent with the observations in ~\cite{ren2019likelihood}. Our motivation is to only leverage the strengths (\ie generating anomaly) and avoid weaknesses (\ie detecting anomaly) of NFs.

\subsection{Multi-Stage Model Training and Inference}

\begin{algorithm}[t]\small
    \caption{\label{algo}\ 
        Multi-stage training
    }
    \begin{tabbing}
        \Inp $\mathcal{ D}_{train} = \{X, Y\}$, number of epoch for training EDL model ($\Phi$) and NFs model ($f$) \\ $T_{\text{EDL}}$, $T_{\text{NFs}}$, thresholds $\tau_{p}$, $\tau_{p}$, $\epsilon$, loss weight $\beta$\= \\
        \Out feature encoder $\mathcal{H}$, EDL classifier $\Phi$.     \= \\
        \X \cm{\textit{Stage 1}: Warmup training} \\
        \XX Train $\mathcal{H}$ and $\Phi$ by minimizing $\mathcal{L}_{\text{MIL}} + \beta \mathcal{L}_{triplet}$. \= \\
        
        \X \cm{\textit{Stage 2}: Freeze $\mathcal{H}$ and train $f$} \\
        \X {\bf for} $t \in [1, T_{\text{NFs}}]$
        {\bf do} 
        \\ 
        \XX Fetch $m$ normal videos, i.e., $\{\mathbf{x}_i\}_{i=1}^{m} \sim \mathcal{ D}_{train}(X,Y=0)$. \\
        \XX Update NFs model $f$ by minimizing $\mathcal{L}_{\text{NF}}$. \\
        \X {\bf end for} \\
        
        \X \cm{\textit{Stage 3}: Freeze $\mathcal{H}$ and $f$, and fine-tune $\Phi$} \\
        \X {\bf for} $t \in [1, T_{\text{EDL}}]$
        {\bf do}  
        \\
        \XX Fetch $m$ normal and anomaly videos from $\mathcal{ D}_{train}$. \\
        \XX Construct the clean positive set $\Omega$ using Eq.~\eqref{eq:selection}. \\
        \XX Augment $\Omega$ by sampling $m$ pseudo anomalies using Eq~\eqref{eq:sampling}, i.e., $\Omega\leftarrow \Omega \cup \mathcal{D}_g$. \\
        
        \XX Update EDL model $\Phi$ by minimizing $\mathcal{L}_{\text{EDL}}$.  \\
        \X {\bf end for}
        
    \end{tabbing}
\end{algorithm}

In this paper, since the NFs is responsible for generating pseudo anomaly instances for the target MIL model, we propose a multi-stage model training scheme. Specifically, we first train a feature encoder $\mathcal{H}(\cdot)$ and an EDL classifier $\Phi(\cdot)$ by minimizing the weighted sum of MIL loss and triplet loss. Then, the feature encoder $\mathcal{H}$ is frozen and the NFs $f(\cdot)$ is trained in an unsupervised way using only normal videos. Finally, the EDL classifier is fine-tuned by using the selected positive instances, the generated pseudo anomaly instances, and the normal videos. Algorithm~\ref{algo} shows the details of the training process. During inference, the NFs does not involve as we use the prediction of EDL classifiers to score samples. The anomaly score is defined as the mean probability of being an anomaly: $\mathbb{E}[p_{+}]=\frac{\alpha_{+}}{\alpha_{0}}$.

\section{Experiments}
Since OpenVAD is newly proposed, no existing evaluation setup is available for this task. To adequately assess the method and inform the practical choice, we develop a benchmark and test several state-of-the-art methods. Through comprehensive comparison on various datasets under the setup, we demonstrate the effectiveness of the proposed framework. 

\subsection{Datasets}
We conduct experiments on three video anomaly detection datasets of different scales.
Videos in these datasets are untrimmed, and only video-level labels are provided during training. The frame-level annotations are used for evaluation purpose. 
Detailed information for these three datasets is as follows:
\begin{itemize}
    \item \textbf{XD Violence}~\cite{wu2020not}: This dataset contains totally 4,754 videos collected from real life movies, online videos, sport streaming, surveillance cameras and CCTVs. There are five types of anomaly events, and each anomaly video may contain one or more types of anomalies, including \textit{Abuse, Riot, Shooting, Fighting, Car accident}, and {\em Explosion}.
    \item \textbf{UCF Crime}~\cite{sultani2018real}: This dataset contains totally 1,900 videos collected from surveillance cameras from variant scenes. There  $13$ different types of violence-related anomaly videos: \textit{Abuse, Burglary, Robbery, Stealing, Shooting, Shoplifting, Assault, Fighting, Arson, Explosion, Arrest, Road Accident}, and \textit{Vandalism}. There are 1,610 training videos (of which 810 are anomaly) and 290 test videos (of which 140 are anomaly). 
    \item \textbf{ShanghaiTech Campus}~\cite{liu2018ano_pred}: This dataset consists videos from 13 scenes with different light conditions and camera angles. Anomaly events are defined as actions that contain sudden motions, such as \textit{chasing} and \textit{brawling}, but there is no anomaly type labels. We follow the split in \cite{zhong2019graph}, in which there are totally 238 training videos (of which 63 are anomaly) and 199 testing (of which 44 are anomaly) videos.
\end{itemize}
\noindent\textbf{Evaluation setup.} 
For the OpenVAD task, we consider a more realistic setting, where testing data contains unseen novel anomalies. To simulate unseen anomaly, we remove one or more types of anomaly from the training data on XD Violence and UCF Crime. We vary the size of anomaly training videos to get comprehensive results under different ratios of unseen anomaly scenarios. For example, in XD-Violence, there are six types of testing anomalies, and we train a model with 1, 2, 3, and 4 types of anomaly videos along with normal videos. We randomly remove a certain number of training anomaly classes and repeat them three times to reduce the evaluation bias of seen anomaly. These evaluation sets reflect our assumption about the most realistic video anomaly detection scenario in an open world. We further test our method with limited number of anomaly videos on the ShanghaiTech dataset. We randomly use a small subset of training anomaly videos. We follow the previous work to use Area under Operating Characteristic curve (AUC-ROC) as the metric on UCF-Crime and ShanghaiTech, and use Area under Precision and Recall curve (AUC-PR) on the XD-Violence dataset. Both AUC-ROC and AUC-PR depict the overall performance using a sliding threshold, and a larger value indicates better performance.

\subsection{Comparison with State-of-the-arts}
In the experiments, we use the pre-trained I3D classifier \cite{carreira2017quo} to extract features for every non-overlapped video clip (each clip is a collection of 16 consecutive frames) in a video. During training, we uniformly sample 200/200/32 (XD-Violence/UCF-Crime/ShanghaiTech) video clips features to create a bag with the same size. 

Our model shares the core design across three datasets and different takes. We use two layers of GCNs and construct the EDL model with two fully connected layers stacked with a ReLU activation function. For the NFs, we use the 5 consecutive IAF~\cite{kingma2016improved}. See Appendix Section B for more implementation details.


\begin{table*}[t!]
    \centering
    \caption{AUC-PR (\%) results on XD-Violence for anomaly frame detection with various number of \textbf{seen anomaly classes}. Methods with (*) in the first column are unsupervised, while the rest are weakly-supervised.}
    %
    \begin{small}
        \begin{sc}
            \begin{tabular}{*{1}{p{3.7cm}|} *{5}{C{1.50cm}}}
                \toprule
                No. seen anomaly & 0 & 1 & 2 & 3 & 4 \\
                \midrule
                OCSVM~\cite{smeureanu2017deep} (*)  & 27.25 & - & - & - & - \\
                Conv-AE~\cite{hasan2016learning}  (*)   & 30.77 & - & - & - & - \\ 
                \midrule
                Wu \etal~\cite{wu2020not}(off-line) & - &  40.67 & 50.34 & 60.53 & 67.77\\
                Wu \etal~\cite{wu2020not}(on-line) & - & 39.13 & 50.20  &58.87 & 64.29 \\
                RTFM \etal~\cite{tian2021weakly} & - & 43.54 & 50.88 & 58.52 & 63.65  \\
                \textbf{Ours} & - & \textbf{45.65} & \textbf{54.65} & \textbf{64.40} & \textbf{69.61} \\
                \bottomrule
            \end{tabular}
            \label{tab:xd}
        \end{sc}
    \end{small}
\end{table*}

\noindent\textbf{Results on Xd-Violence.} Table~\ref{tab:xd} shows the AUC-PR scores in detecting anomaly frames in settings where 6 types anomaly activities appear in testing, while only 1, 2, 3, or 4 types of them are seen during training. Our method outperforms both unsupervised and weakly supervised ones in all cases. In particular, our method achieves a significant performance gain when very few anomalies are seen, which clearly justifies its capability in the open world setting. Hasan \etal~\cite{hasan2016learning} and OCSVM~\cite{smeureanu2017deep} do not rely on anomaly videos compared to other methods. Hasan \etal surpasses the OCSVM by around 3\% AUC-PR scores. However, the weakly supervised methods perform much better when there are anomaly videos available for training. Even in the extreme case when only 1 out of 6 types of anomalies are seen, the weakly supervised methods can outperform the unsupervised ones by more than 10\%. Moreover, more anomaly videos for training further boost the performance of weakly supervised methods. Our method also achieves better a AUC-ROC score than other weakly supervised methods, indicating that our method makes less false positive predictions (See the AUC-ROC score in Appendix Section D).       
\begin{table*}[ht]
    \centering
    \caption{AUC-ROC(\%) results on UCF-Crime for video anomaly detection with various number of \textbf{seen anomaly classes}.
    Methods with (*) in the first column are unsupervised, while the rest are weakly-supervised.}
    %
    \begin{small}
        \begin{sc}
            \begin{tabular}{*{1}{p{3.7cm}|} *{5}{C{1.50cm}}}
                \toprule
                Num. anomaly    & 0 & 1 & 3 & 6 & 9 \\
                \midrule
                Sohrab\etal~\cite{sohrab2018subspace}(*) &
                58.50 & - & - & - & - \\
                Lu \etal~\cite{lu2013abnormal}(*) & 
                65.51 & - & - & - &- \\
                BODS~\cite{wang2019gods}(*) & 
                68.26 & - & - & - &- \\
                GODS~\cite{wang2019gods}(*) & 
                70.46 & - & - & - &- \\
                Conv-AE~\cite{hasan2016learning}  (*) & 
                50.60 & - & - & - &- \\
                \midrule
                Wu \etal\cite{wu2020not}(off-line) & -& 73.22 & 75.15 &  78.46   &  79.96 \\
                Wu \etal\cite{wu2020not}(on-line) & -& 73.78 & 74.64 &  77.84 &  79.11 \\
                RTFM \etal~\cite{tian2021weakly} & - & 75.91 & 76.98 &77.68 & 79.55 \\
                \textbf{Ours} &  - & \textbf{76.73} & \textbf{77.78} & \textbf{78.82} & \textbf{80.14} \\
                \bottomrule
            \end{tabular}
            \label{tab:ucf}
        \end{sc}
    \end{small}
\end{table*}

\begin{table*}[ht]
    \centering
    \caption{AUC-ROC(\%) results on ShanghaiTech for video anomaly detection with various number of \textbf{seen anomaly videos}. Methods with (*) in the first column are unsupervised, while the rest are weakly-supervised.}
    %
    \begin{small}
        \begin{sc}
            \begin{tabular}{*{1}{p{4cm}|} *{5}{C{1.50cm}}}
                \toprule
                Num.seen anomaly & 0 & 5 & 10 & 15 & 25\\
                \midrule
                Frame-Pred~\cite{liu2018future}(*)& 73.40  & - &  - &  -  & - \\
                Mem-AE~\cite{gong2019memorizing} (*)& 71.20  & - &  - &  - & - \\
                MNAD~\cite{gong2019memorizing} (*)& 70.50  & - &  - &  - & - \\
                VEC~\cite{yu2020cloze}(*) & 74.80 & - &  - &  - & - \\
                Conv-AE~\cite{hasan2016learning}(*) &  60.85 & - &  - &  - & - \\
                \midrule
                Wu \etal\cite{wu2020not}(off-line) & -& 65.83 & 81.54 &  83.47 & 88.81\\
                Wu \etal\cite{wu2020not}(on-line) & -& 66.27 & 81.03 &  82.42  & 88.61 \\
                RTFM \etal~\cite{tian2021weakly} & - & 70.59 & 83.42 & 81.50 & 86.33 \\
                \textbf{Ours} & -  &  \textbf{80.40} & \textbf{88.24}  & \textbf{85.58}  & \textbf{93.99}  \\
                \bottomrule
            \end{tabular}
            \label{tab:shanghai}
        \end{sc}
    \end{small}
\end{table*}

\noindent\textbf{Results on UCF-Crime.} Table~\ref{tab:ucf} shows that AUC-ROC scores in detecting anomaly frames when 13 types of the anomaly appears during testing, only 1, 3, 6, and 9 types of these anomaly are seen. Results in Table~\ref{tab:ucf} show that our method consistently outperforms all baselines under all training settings. To our surprise, all weakly supervised methods can achieve relatively good performance even when a small subset of the anomaly is seen. It could be caused that most anomalous behaviours are related to human activities, and there is somehow high correlation among various anomalies, \ie \textit{Abuse}, \textit{Assault}, and \textit{Vandalism}. RTFM \etal~\cite{tian2021weakly} achieves better performance than Wu \etal~\cite{wu2020not} when fewer anomaly classes are seen, which may be owing to its feature magnitude component.  Nevertheless, our method achieves the most stable performance under various training setups, implying that our proposed framework is robust by learning a compact decision boundary. 

\noindent\textbf{Results on ShanghaiTech. } Table~\ref{tab:shanghai} shows the AUC-ROC under an imbalance scenario: only a few anomalies are available for training. Results indicate that our method performs better in the challenging regime when only 25 or fewer anomaly videos are available for training. In that case, our method outperforms both RTFM~\cite{tian2021weakly} and Wu \etal~\cite{wu2020not} by a large margin. In most cases, the performance advantage is more significant when fewer videos are available. This clearly indicates the effectiveness of the proposed method.  

\subsection{Ablation Study}
We conduct ablation study to validate the proposed three components in our framework, \ie (i) Triplet Loss defined in Eq.~\eqref{eq:triplet}, (ii) Evidence Criteria defined in Eq.~\eqref{eq:selection}, and (iii) Pseudo Anomaly generation. Specifically, we incrementally remove/replace them and compare the performance on XD-Violence dataset. When removing the Evidence Criteria, we set the $\Omega$ to include all samples. To evaluate the quality of pseudo anomalies, we replace it with 
Gaussian noise.
\begin{table}[!tb]
    \caption{Ablation study results (\%) on XD-Violence dataset.}
    \centering
    \scriptsize
    \label{tab:ablation}
    \begin{tabular}{c|c|c|cccc}
        \toprule
        \multicolumn{1}{c}{\multirow{2}*{Triplet Loss}} & \multicolumn{1}{|c}{\multirow{2}*{Evidence Criteria}} &
        \multicolumn{1}{|c}{\multirow{2}*{Pseudo Anomaly}} &
        \multicolumn{2}{|c}{1 seen anomaly} &
        \multicolumn{2}{c}{4 seen anomaly} \\
        \cmidrule(r){4-5} \cmidrule(r){6-7} 
        & & & AUC-PR & AUC-ROC & AUC-PR & AUC-ROC \\
        \midrule
        $\surd$  &  $\surd$ & $\surd$ & \textbf{45.65}	& \textbf{72.50} &  \textbf{69.61}	& \textbf{88.25}\\
                 &  $\surd$ & $\surd$ & 45.26 & 61.30 & 69.14 & 83.57\\
                 &   & $\surd$ & 41.79  &  67.07 & 68.82 & 84.78\\
                &   &  &  40.94 & 66.83 & 66.61 & 84.09\\
        \bottomrule
    \end{tabular}
\end{table}

Table~\ref{tab:ablation} validates the contribution of each proposed technique. It shows that the combination of the three components achieves the best AUC-PR score and AUC-ROC score, indicating that all these three components positively contribute to the performance of our framework. Notably, the triplet loss contributes most to the performance gain. This demonstrates the important role of instance representation learning for video anomaly detection. Besides, we note that the evidence criteria produce better AUC-PR scores while worse AUC-ROC scores. This can be explained that the evidence criteria drive the decision boundary toward the positive side in feature space as it filters out the noisy labels. As a result, the false positive predictions are less while the false negative predictions might increase. Lastly, once the pseudo anomaly generation is applied, the performances of both AUC-PR and AUC-ROC are improved.

\section{Conclusion}
In this paper, we present a new approach for the OpenVAD problem under weak supervision, which is a highly challenging task by previous unsupervised and weakly supervised methods when being used alone. To address the unique challenges, the proposed framework integrates unsupervised learning (\ie NFs) and weakly supervised learning (\ie MIL) in novel ways to benefit the advantages of both while not suffering from their individual limitations. Different from existing  unsupervised methods, the proposed approach makes use of any available anomaly videos without expensive labelling costs. Unlike other weakly supervised methods, our method can detect any type of seen and unseen anomalies in an open world. The OpenVAD scenario is the most realistic scenario in real-world applications, as the anomaly events are 
difficult to be fully 
modeled. 
~\\[6pt]
\textbf{Acknowledgements:}
 This research is supported by Office of Naval Research (ONR) grant N00014-18-1-2875 and the Army Research Office (ARO) grant W911NF-21-1-0236. The views and conclusions in this document are those of the authors and should not be interpreted as representing the official policies, either expressed or implied, of the ONR, the ARO, or the U.S. Government.

\clearpage
%
%
\bibliographystyle{splncs04}
\bibliography{egbib}

\newpage
\appendix
\title{Towards Open Set Video Anomaly Detection\\
Supplementary Material} 
\titlerunning{Towards Open Set Video Anomaly Detection}
%
\author{Yuansheng Zhu
\and
Wentao Bao
\and
Qi Yu
}
\authorrunning{Zhu, Bao, and Yu}
%
\institute{
Rochester Institute of Technology \\
\email{\{yz7008, wb6219 and qi.yu\}@rit.edu}}
\maketitle

The Appendix is organized as follows. In Section~\ref{app:notations}, we summarize the major notions used in the paper. In Section~\ref{app:impdetail}, we provide the implementation details. In Section~\ref{app:additional}, we provide additional results. In Section~\ref{app:visual}, we show some qualitative examples.

\section{Notations}\label{app:notations}
The main notations are divided into four major types: Data, Model, Loss, and Hyperparameters, and summarized in Table~\ref{tab:notation}.
\begin{table}[ht]
    \caption{Summary of notations\label{tab:notation}}
    \centering
    \scriptsize
    \begin{tabular}{*{1}{p{3cm}}| *{1}{C{8cm}}}
        \toprule
        {\bf Type}&   \textbf{Notation} \\
        \midrule
        \multirow{2}{*}{\textbf{Data}} &  Bag feature $X$, Bag label $Y$, instance feature 
        $\mathbf{x}$, instance label $y$ \\
        \cmidrule{2-2}
        & Adjacent matrix $A$  \\
        \midrule
        \multirow{3}{*}{\textbf{Model}} & GCNs $\mathcal{H}(\cdot)$\\
        \cmidrule{2-2}
        & EDL $\Phi (\cdot)$\\
        \cmidrule{2-2}
        & NFs $f(\cdot)$\\
        \midrule
        \multirow{3}{*}{\textbf{Loss}} & Triplet loss $\mathcal{L}_{triplet}$\\
        \cmidrule{2-2}
        & MIL loss $\mathcal{L}_{\text{MIL}}$\\
        \cmidrule{2-2}
        & NFs loss $\mathcal{L}_{NF}$\\
        \midrule
        \multirow{3}{*}{\textbf{Hyperparameters}} & Loss weight $\beta$\\
        \cmidrule{2-2}
        & Triplet loss margin $m$ \\
        \cmidrule{2-2}
        & thresholds $\tau_{u}$, $\tau_{p}$, $\epsilon$ for constructing $\Omega$ \\
        \bottomrule
    \end{tabular}
\end{table}

\section{Implementation Details}\label{app:impdetail}

The hyperparameters are chosen as follows: $m$ is set as $0.3$ across three datasets, and $\beta$ is set as $0.001$, $0.0001$, $0.0001$ for XD-Violence, UCF-Crime, and ShanghaiTech, respectively. To be adaptive during the training process, $\tau_{p}$, $\tau_{u}$, and $\epsilon$ are chosen based on the $i$-th largest value in a candidate pool during every iteration. Generally, $\tau_{p}$ and $\tau_{u}$ are set to make $\Omega$ retain a moderate portion of instances in every bag, and $\epsilon$ is set to make the pseudo anomalies with low probability density. In practice, on XD-Violence, ShanghaiTech, UCF-Crime, $\tau_{p}$ is set as the $50$-{th}, $30$-{th}, $3$-{rd} largest $p_{+}$, and $\tau_{u}$ is set as the the $150$-{th}, $150$-{th}, and $24$-{th} largest $\alpha_{+}$ in a bag. $\epsilon$ is set as the $4750$-{th} largest $p\left(\tilde{\mathbf{x}}|y=0\right)$ in a pseudo anomaly pool of size $5000$. We gradually perform sample selection, \ie increasing $\tau_{p}$ from smallest to the assigned value during a warmup stage ($\Omega$ evolves from all instances to the most confident clean subset). We perform early stopping to avoid overfitting whenever needed. We optimize the model via the Adam optimizer equipped with cosine annealing learning rate scheduling. We use Python 3.9.7 and PyTorch 1.10.0 to build the test platform, running it on NVIDIA RTX A6000 GPUs. Whenever public results are available, we directly use them for comparison. 

\section{Additional Results}\label{app:additional}
In this section, we present more experimental results along with an additional ablation study to further justify the key components of the proposed framework. 

\subsection{AUC-ROC on XD-Violence}
We show the AUC-PR scores on the XD-Violence in the main paper because it is used in previous works~\cite{wu2020not,tian2021weakly} for this dataset. In combination with the AUC-PR, we provide the AUC-ROC scores in Table~\ref{tab:xd-roc}, which are collected under the same setting. It can be seen that our method achieves the highest AUC-ROC scores among the weakly supervised methods under all settings, and the conclusion using two metrics are consistent.


\begin{table*}[ht]
    \centering
    \caption{AUC-ROC (\%) results on XD-Violence for anomaly frame detection with various number of \textbf{seen anomaly classes}.}
    \vspace{2mm}
    \begin{small}
        \begin{sc}
            \begin{tabular}{*{1}{p{3.7cm}|} *{4}{C{1.50cm}}}
                \toprule
                No. seen anomaly  & 1 & 2 & 3 & 4 \\
                \midrule
                Wu \etal~\cite{wu2020not}(off-line) &  67.05 & 71.88 & 73.06 & 85.32\\
                Wu \etal~\cite{wu2020not}(on-line) & 66.13 & 72.32  & 72.49 & 83.49 \\
                RTFM~\cite{tian2021weakly} & 66.54 & 70.78 & 76.70 & 82.41 \\
                \textbf{Ours} & \textbf{72.50} & \textbf{77.51} & \textbf{84.57} & \textbf{88.25} \\
                \bottomrule
            \end{tabular}
            \label{tab:xd-roc}
        \end{sc}
    \end{small}
\vspace{-2mm}
\end{table*}

\begin{table}[ht]
    \centering
    \caption{Ablation study results for anomaly frame detection on XD-Violence in close-world Setting (NUM ANOMALY=ALL).}
    \begin{small}
        \begin{sc}
            \begin{tabular}{*{1}{p{3.7cm}|} *{2}{C{2cm}}}
                \toprule
                  & AUC-PR & AUC-ROC \\
                \midrule
                Wu \etal~\cite{wu2020not}(off-line) & 75.80
                &  93.07 \\
                Wu \etal~\cite{wu2020not}(on-line) & 72.92 & 92.02\\
                RTFM~\cite{tian2021weakly} & 69.40
                & 88.09\\
                \midrule
                NFs(w/o Trip) & 52.10 & 77.40 \\
                NFs & 73.13 & 89.77 \\
                \midrule
                Ours(w/o GCNs) & 69.39 & 89.14 \\
                Ours(w top-{k}) & 77.43 & 92.66\\
                \textbf{Ours} & 77.91 &  93.23\\
                \bottomrule
            \end{tabular}
            \label{tab:xd-full}
        \end{sc}
    \end{small}
\end{table}


\subsection{Additional Ablation Study}
In Table~\ref{tab:xd-full}, we provide additional ablation study results on the XD-Violence under the close set setting. For the ablation study, the NFs and NFs (w/o Triplet) denote using the NFs to score a sample during testing, and NFs (w/o Triplet) mean that we remove the Triplet loss. To explore the impact of feature encoder, we replace the GCNs with two FC layers, denoted as the Ours (w/o GCNs). Finally, we provide results of the top-$k$ selection by setting $\Omega =\{{\bf x}_{i}|p_{i}>\tau_{p}\}$.

Table~\ref{tab:xd-full} shows the results of two weakly supervised baselines, NFs, and ours, under a close set setting. To use the NFs for anomaly detection, we leverage its density estimation capability to score a sample, \ie a sample with low density is considered to be likely to be an anomaly, similar to the usage of NFs with Cho \etal~\cite{cho2022unsupervised}. Results show that the triplet loss contributes a lot to the performance of NFs, proving its important role in facilitating the learning process of NFs (See the NFs vs NFs w/o Trip). Besides, when the GCNs is equipped with triplet loss for representation learning, NFs can achieve comparable performance with the Wu \etal~\cite{wu2020not} and RTFM~\cite{tian2021weakly}. Nevertheless, our approach outperforms the NFs by a large margin, justifying the advantage of our usage of NFs over the previous use (pseudo anomaly generation vs density estimation). 

Results also show that the choice of feature encoder significantly impacts the anomaly detector; the performance drops a lot when replacing the GCNs with FC layers (See ours vs ours w/o GCNs). 
We also compare our evidence-based instance selection with the top-{$k$} strategy. Based upon top-{$k$}, which solely uses the predicted probability $p_{+}$ to perform selection, our instance selection method adds the evidence $\alpha_{+}$ to improve its robustness. The relation between $p_{+}$ and $\alpha_{+}$ is determined by $\mathbb{E}[p_{+}]=\frac{\alpha_{+}}{\alpha_{+}+\alpha_{-}}$, where $p_{+}$ and $\alpha_{+}$ denote the probability of being positive and evidence of supporting a positive prediction, respectively. After acquiring the evidence, we use $\tau_{\alpha}$ to filter out samples that are likely the false anomaly. Comparison between ours and top-$k$ shows that adding the evidence could improve the robustness of the latter. We remark that existing literature also uses $u$ to estimate the predictive uncertainty rather than using $\alpha_{+}$. However, using $\alpha_{+}$ achieves similar, but superior, effect compared with $u$ because $u$ is upper bounded by $\frac{2}{\tau_{\alpha}}$: $u=\frac{2}{\alpha_{+}+\alpha_{-}}<\frac{2}{\alpha_{+}}\leq\frac{2}{\tau_{\alpha}}$. Among the samples with low $u$, using $\alpha_{+}$ would prefer the desired confident anomaly ones.




\section{Qualitative Results}\label{app:visual}

\begin{figure*}[htpb]
\centering
\subfloat[Shooting]{\includegraphics[width=0.90\linewidth]{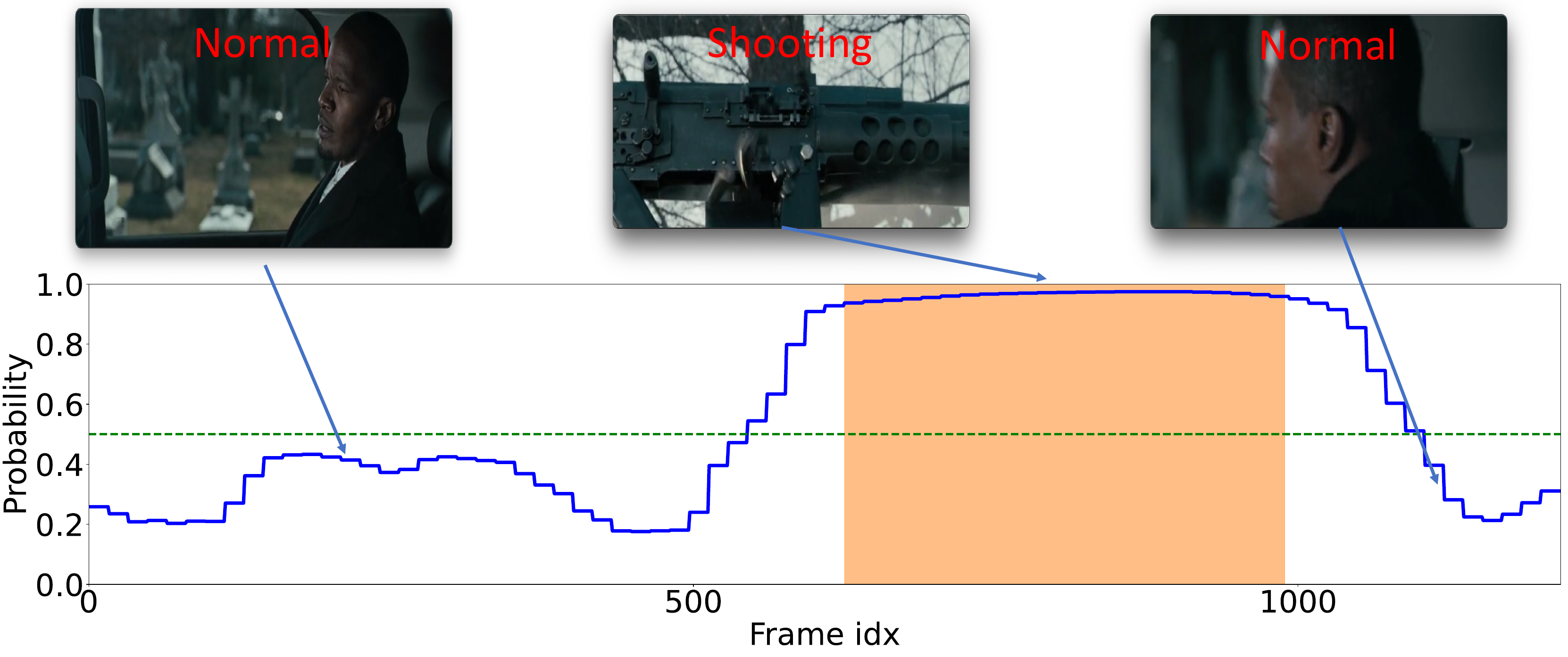}}\quad \\
\subfloat[Normal]{\includegraphics[width= 0.90\linewidth]{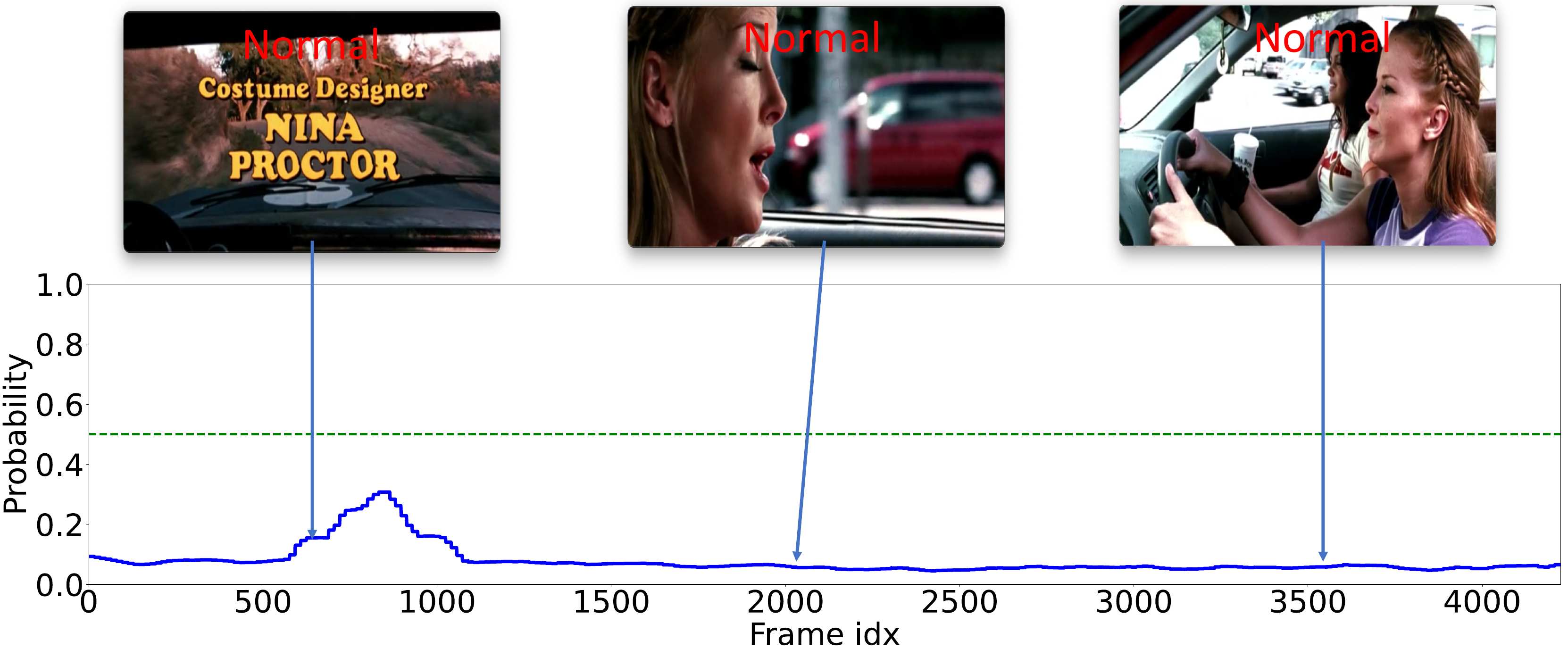}}\quad\\
\caption{Visualized results on XD-Violence for \textbf{seen} anomaly frame detection in (a) a {\em Shooting} video and (b) a normal video. The top row in each example shows raw frames from the video, and the bottom row shows the predicted anomaly score (\textcolor{blue}{blue curve}) with ground-truth anomaly regions (\textcolor{orange}{orange window}). Model is trained with {\em Fighting, Shooting, Abuse, Explosion} and normal videos. {\em Riot} and {\em Car accident} are set aside as unseen anomalies.}
\label{fig:visual_example_seen}
\end{figure*}


\begin{figure*}[htpb]
\centering
\subfloat[Car accident]{\includegraphics[width= 0.90\linewidth]{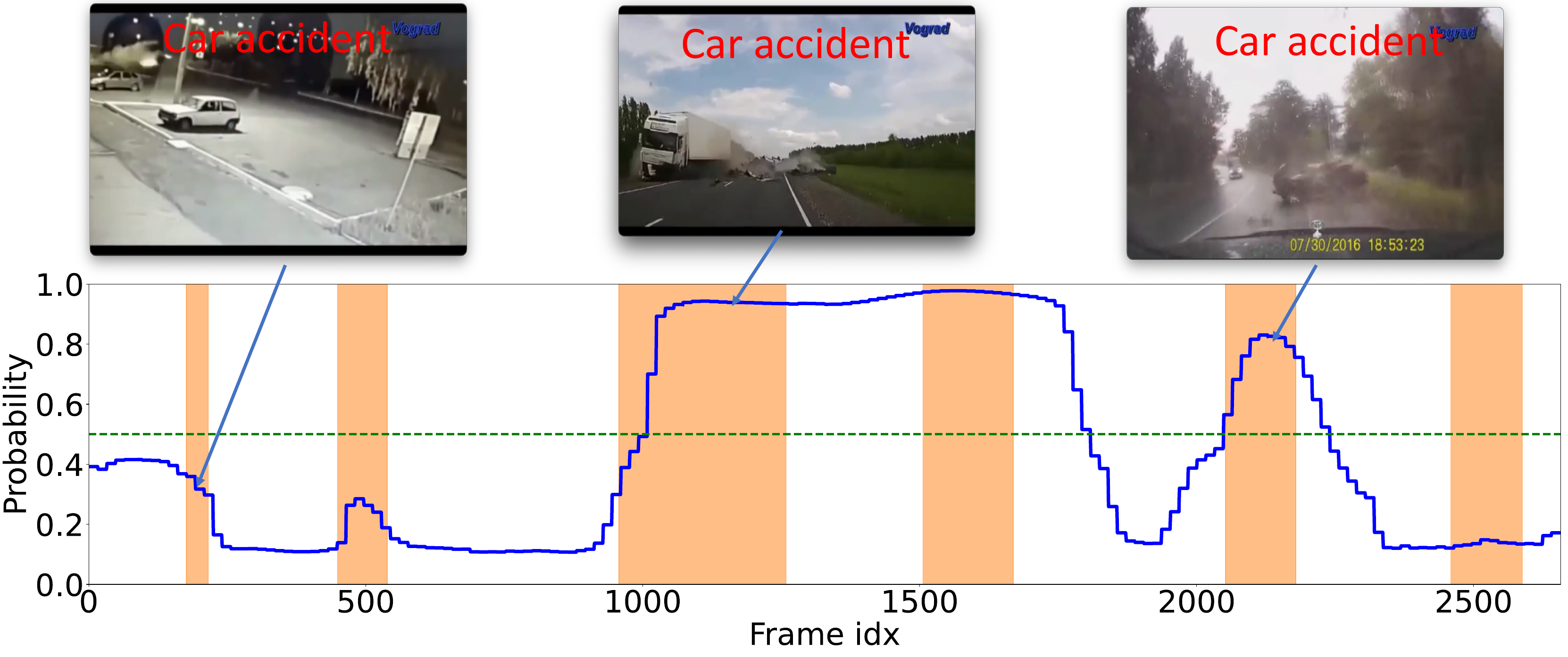}}\quad \\
\subfloat[Riot]{\includegraphics[width= 0.90\linewidth]{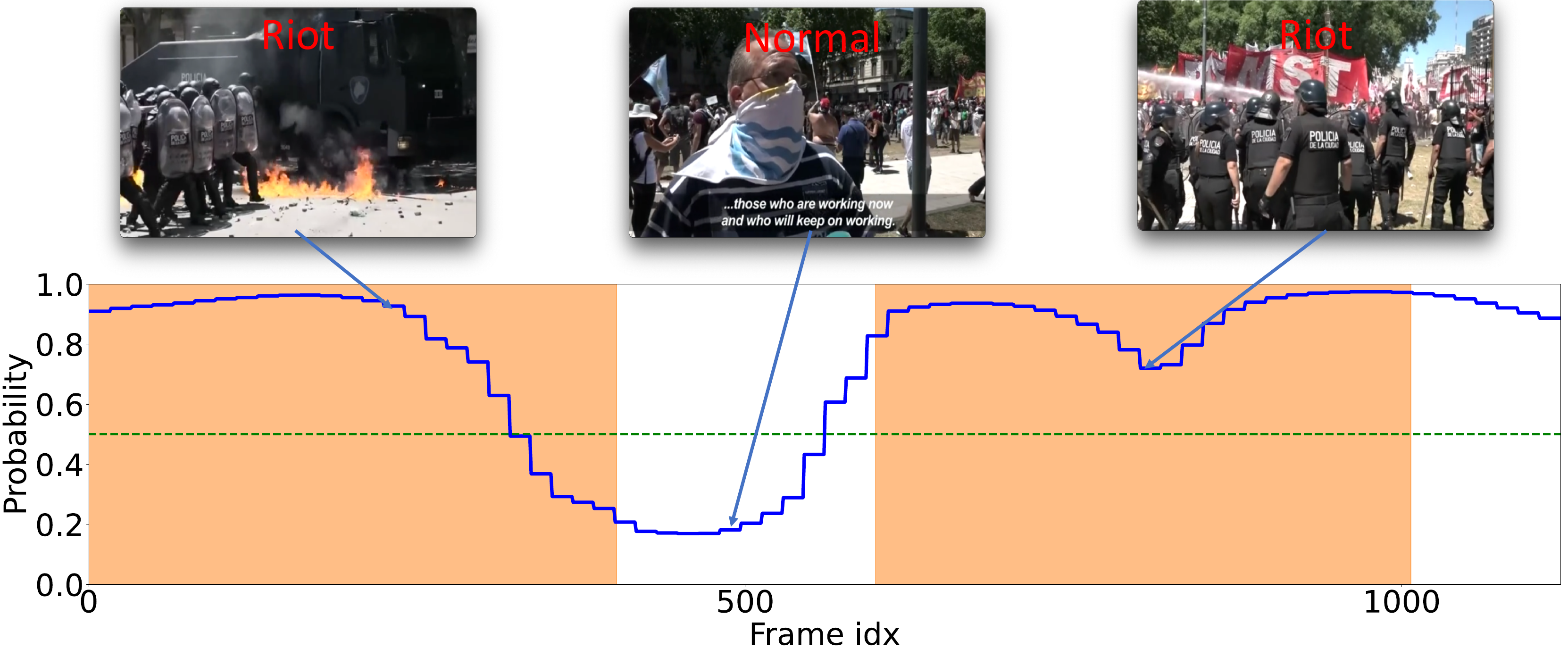}}\quad\\
\caption{Visualized results on XD-Violence for \textbf{unseen} anomaly frame detection in (a) a {\em Car Accident} video and (b) a {\em Riot} video. The top row in each example shows raw frames from the video, and the bottom row shows the predicted anomaly score (\textcolor{blue}{blue curve}) with ground-truth anomaly regions (\textcolor{orange}{orange window}). Model is trained with {\em Fighting, shooting, Abuse, Explosion} and Normal videos. {\em Riot} and {\em Car Accident} are set aside as unseen anomalies.}
\label{fig:visual_example_useen}
\end{figure*}


We plot the results of a model trained with $4$ types of anomalies on the XD-Violence dataset. Figure~\ref{fig:visual_example_seen} shows that our model fully captures the anomaly region (\ie  {\em Shooting}) as they have been seen during training. For the unseen anomaly frames, which are more challenging, Figure~\ref{fig:visual_example_useen} shows that our model performs well on detecting them, especially the {\em Riot}. Our model misses in detecting some {\em Car accident} events as they last briefly. We also note that our model gives relatively high anomaly scores to some normal frames in anomaly videos, but the margin between anomaly and normal ones is still noticeable. This can be explained that these frames show a sign of violence and are ambiguous, while they are labelled as normal by the human annotator. These observations validate the effectiveness of our model for the proposed OpenVAD task, \ie detecting arbitrary anomalies.

\end{document}